\documentclass[11pt,a4paper]{article}
\usepackage[hyperref]{acl2019}
\usepackage{times}
\usepackage{latexsym}

\usepackage{graphicx}
\usepackage{multirow}
\usepackage{amsmath}
\usepackage{booktabs}
\usepackage{caption}
\usepackage{subcaption}
\usepackage{siunitx}

\usepackage{tabularx}
\usepackage{changepage}

\usepackage{booktabs}

\usepackage{url}

\aclfinalcopy %
\def\aclpaperid{2306} %

\newcommand{\eg}[0]{e.g.\ }

\title{Cross-Domain Generalization of Neural Constituency Parsers}

\author{
Daniel Fried\thanks{\enskip Equal contribution.} \qquad Nikita Kitaev\footnotemark[1] \qquad Dan Klein \\
Computer Science Division \\
University of California, Berkeley \\
{\tt \{dfried,kitaev,klein\}@cs.berkeley.edu}
}

\date{}

\begin{document}
\maketitle
\begin{abstract}

Neural parsers obtain state-of-the-art results on benchmark treebanks for constituency parsing---but to what degree do they generalize to other domains? We present three results about the generalization of neural parsers in a zero-shot setting: training on trees from one corpus and evaluating on out-of-domain corpora. First, neural and non-neural parsers generalize comparably to new domains. Second, incorporating pre-trained encoder representations into neural parsers substantially improves their performance across all domains, but does not give a larger relative improvement for out-of-domain treebanks. Finally, despite the rich input representations they learn, neural parsers still benefit from structured output prediction of output trees, yielding higher exact match accuracy and stronger generalization both to larger text spans and to out-of-domain corpora. We analyze generalization on English and Chinese corpora, and in the process obtain state-of-the-art parsing results for the Brown, Genia, and English Web treebanks.

\end{abstract}

\begin{table*}[t]
    \centering
    \footnotesize
    \begin{tabular}{r|cc|cc|cc|cc}
    \toprule
 &  \multicolumn{2}{c|}{Berkeley} &  \multicolumn{2}{c|}{BLLIP} &  \multicolumn{2}{c|}{In-Order} &  \multicolumn{2}{c}{Chart} \\
 &  F1 & $\Delta$ Err. &  F1 & $\Delta$ Err. &  F1 & $\Delta$ Err. &  F1 & $\Delta$ Err. \\
 \midrule
WSJ Test & 90.06 & \phantom{00}+0.0\% & 91.48 & \phantom{00}+0.0\% & 91.47 & \phantom{00}+0.0\% & 93.27 & \phantom{00}+0.0\% \\
\midrule
Brown All & 84.64 & \phantom{0}+54.5\% & 85.89 & \phantom{0}+65.6\% & 85.60 & \phantom{0}+68.9\% & 88.04 & \phantom{0}+77.7\% \\
Genia All & 79.11 & +110.2\% & 79.63 & +139.1\% & 80.31 & +130.9\% & 82.68 & +157.4\% \\
EWT All & 77.38 & +127.6\% & 79.91 & +135.8\% & 79.07 & +145.4\% & 82.22 & +164.2\% \\
\bottomrule
    \end{tabular}
    \caption{\label{tab:neural_ood_english}Performance and relative increase in error (both given by F1) on English corpora as parsers are evaluated out-of-domain, relative to performance on the in-domain WSJ Test set. Improved performance on WSJ Test translates to improved performance out-of-domain. The two parsers with similar absolute performance on WSJ (BLLIP and In-Order) have comparable generalization out-of-domain, despite one being neural and one non-neural.}
    \vspace{-1em}
\end{table*}

\section{Introduction}

Neural constituency parsers have obtained increasingly high performance when measured by F1 scores on in-domain benchmarks, such as the Wall Street Journal (WSJ) \cite{marcus1993building} and Penn Chinese Treebank (CTB) \cite{xue2005penn}.
However, in order to construct systems useful for cross-domain NLP, we seek parsers that generalize well to domains other than the ones they were trained on.
While classical, non-neural parsers are known to perform better in their training domains than on out-of-domain corpora, their out-of-domain performance degrades in well-understood ways \cite{gildea2001corpus,petrov-klein:2007:main}, and improvements in performance on in-domain treebanks still transfer to out-of-domain improvements \cite{mcclosky2006reranking}.

Is the success of neural constituency parsers (\citealt{henderson2004discriminative,vinyals2015grammar,dyer2016rnng,cross2016span,choe2016parsing,stern2017minimal,liu2017inorder,kitaev2018constituency}, \emph{inter alia}) similarly transferable to out-of-domain treebanks?
In this work, we focus on \emph{zero-shot generalization}: training parsers on a single treebank (\eg WSJ) %
and evaluating on a range of broad-coverage, out-of-domain treebanks (\eg Brown \cite{francis1979manual}, Genia  \cite{tateisi2005genia}, the English Web Treebank \cite{petrov2012overview}). We ask three questions about zero-shot generalization properties of state-of-the-art neural constituency parsers: %

First, \emph{do non-neural parsers have better out-of-domain generalization than neural parsers?}
We might expect neural systems to generalize poorly because they are highly-parameterized, and may overfit to their training domain. %
We find that \textbf{neural and non-neural parsers generalize similarly}, and, encouragingly,
improvements on in-domain treebanks still transfer to out-of-domain.%

Second, \emph{does pre-training particularly improve out-of-domain performance}, or  does it just generally improve test accuracies?
Neural parsers incorporate rich representations of language 
that can easily be pre-trained on large unlabeled corpora \cite{ling2015simple,peters2018elmo,devlin2018bert} and
improve accuracies in new domains \cite{joshi2018extending}.
Past work has shown that 
lexical supervision on an out-of-domain treebank can substantially improve parser performance \cite{rimell2009porting}.
Similarly, we might expect pre-trained language representations to give the largest improvements 
on out-of-domain treebanks, by providing 
representations of language disparate from the training domains.
Surprisingly, however,
we find that \textbf{pre-trained representations give similar error reductions across domains}.

Finally, \emph{how much does structured prediction help neural parsers?} 
While neural models with rich modeling of syntactic structure have obtained strong performance on parsing \cite{dyer2016rnng,liu2017inorder} and a range of related tasks \cite{kuncoro2018dependencies,hale2018syntax}, recent  
neural parsers obtain state-of-the-art F1 on benchmark datasets using rich input encoders without any explicit modeling of correlations in output structure \cite{shen2018straight,kitaev2018constituency}. Does structural modeling still improve parsing performance even with these strong encoder representations? We find that, yes, while structured and unstructured neural models (using the same encoder representations) obtain similar F1 on in-domain datasets, \textbf{the structured model  typically generalizes better to longer spans and out-of-domain treebanks, and has higher exact match accuracies in all domains}. %

\section{Experimental setup}
\label{sec:setup}
We compare the generalization of strong non-neural parsers against recent state-of-the-art neural parsers on English and Chinese corpora.%

\paragraph{Non-neural models}
We use publicly released code and models for the Berkeley Parser \cite{petrov-klein:2007:main} and BLLIP Parser \cite{charniak2000maximum,charniak2005coarse} for English; and ZPar \cite{zhang2011syntactic} for Chinese.

\paragraph{Neural models}
We use two state-of-the-art neural models: the Chart model of \citet{kitaev2018constituency}, and In-Order shift-reduce model of \citet{liu2017inorder}. These parsers differ in their modeling both of input sentences and output structures. The Chart model uses a self-attentive encoder over the input sentence, and does not explicitly model output structure correlations, predicting tree span labels independently conditioned on the encoded input.\footnote{The only joint constraint on span predictions is to ensure they constitute a valid tree.} The In-Order shift-reduce model of \citet{liu2017inorder} uses a simpler LSTM-based encoding of the input sentence but a decoder that explicitly conditions on previously constructed structure of the output tree, obtaining the best performance among similarly structured models \cite{dyer2016rnng,kuncoro2016recurrent}. 

The In-Order model conditions on predicted part-of-speech tags; we use tags predicted by the Stanford tagger (following the setup of \citet{cross2016span}). At test time, we use Viterbi decoding for the Chart model and beam search with beam size 10 for the In-Order model.

To control for randomness in the training procedure of the neural parsers, all scores reported in the remainder of the paper  for the Chart and In-Order parsers 
are averaged across five copies of each model trained from separate random initializations.

\paragraph{Corpora}
The English parsers are trained on the WSJ training section of the Penn Treebank. We perform in-domain evaluation of these parsers on the WSJ test section, and out-of-domain evaluation using the Brown, Genia, and English Web Treebank (EWT) corpora.
For analysis and comparisons within parsers, we evaluate on the entirety of each out-of-domain treebank; for final results and comparison to past work we use the same testing splits as the past work.

The Chinese parsers are trained on the training section of the Penn Chinese Treebank (CTB) v5.1 \cite{xue2005penn}, consisting primarily of newswire. For out-of-domain evaluation on Chinese, we use treebank domains introduced in CTB versions 7 and 8: broadcast conversations (B.\ Conv), broadcast news (B.\ News), web discussion forums (Forums) and weblogs (Blogs).

\section{How well do neural parsers generalize?}
\label{sec:neural_generalization}

Table~\ref{tab:neural_ood_english} compares the generalization performance of the English parsers, both non-neural (Berkeley, BLLIP) and neural (Chart, In-Order). None of these parsers use additional data beyond the WSJ training section of the PTB: we use the version of the BLLIP parser without self-training on unlabeled data, and use the In-Order parser without external pre-trained word embeddings. Across all parsers, higher performance on the WSJ Test set corresponds to higher performance on each out-of-domain corpus, showing that the findings of \citet{mcclosky2006reranking} extend to recent neural parsers. In particular, the Chart parser has highest performance in all four domains.

The $\Delta$ Err.\ column shows the \emph{generalization gap} for each parser on each corpus: the parser's relative increase in error (with error defined by $100-\textsc{F1}$) from the WSJ Test set (lower values are better). Improved performance on the WSJ Test set corresponds to increased generalization gaps, indicating that to some extent parser improvements on WSJ have come at the expense of out-of-domain generalization. However, the two parsers with similar absolute performance on WSJ---the BLLIP parser and In-Order parser---have comparable generalization gaps, despite one being neural and one non-neural.

\begin{table}[t]
\centering
\footnotesize
\begin{tabular}{r|cc|cc}
\toprule
 &  \multicolumn{2}{c|}{ZPar} &  \multicolumn{2}{c}{In-Order} \\
 &  F1 & $\Delta$ Err. &  F1 & $\Delta$ Err. \\
\midrule
CTB Test & 83.01 & \phantom{0}+0.0\% & 83.67 & \phantom{0}+0.0\%\\
\midrule
B.\ News & 77.22 & +34.1\% & 77.83 & +35.8\%\\
Forums & 74.31 & +51.2\% & 75.71 & +48.7\%\\
Blogs & 73.90 & +53.6\% & 74.74 & +54.7\%\\
B.\ Conv. & 66.70 & +96.0\% & 67.69 & +97.9\%\\
\bottomrule
\end{tabular}
    \caption{\label{tab:neural_ood_chinese}Performance on Chinese corpora and increase in error (relative to the CTB test set) as parsers are evaluated out-of-domain. The non-neural (ZPar) and neural (In-Order) parser generalize similarly.}
    \vspace{-1.5em}
\end{table}

Table~\ref{tab:neural_ood_chinese} shows results for ZPar and the In-Order parser on the Chinese treebanks, with $\Delta$ Err.\ computed relative to the in-domain CTB  Test set. As with the English parsers and treebanks, increased performance on the in-domain test set corresponds to improvements on the out-of-domain treebanks (although these differences are small enough that this result is less conclusive than for English). In addition, as with English, we observe similar generalization performance of the non-neural and neural parsers across the out-of-domain treebanks.

\section{How much do pretrained representations help out-of-domain?}
Pre-trained word representations have been shown to increase in-domain parsing accuracies. Additionally, \citet{joshi2018extending} showed that these representations (in their case, from ELMo, \citealt{peters2018elmo}) allow a parser to transfer well across domains.
We analyze whether pre-trained representations provide a greater benefit in-domain or out-of-domain, by comparing %
relative performance improvements on in-domain and out-of-domain treebanks when augmenting the neural parsers with pre-trained language representations. We evaluate non-contextual word embeddings produced by structured skip-gram \cite{ling2015simple}, as well as the current state-of-the-art contextual representations from BERT \cite{devlin2018bert}. 

\begin{table}[]
    \centering
    \footnotesize
    \scalebox{0.94}{
    \begin{tabular}{r|c|cc|cc}
    \toprule
 &  In-Order &  \multicolumn{2}{c|}{+Embeddings} &  \multicolumn{2}{c}{+BERT} \\
 &  F1 &  F1 & $\Delta$ Err. &  F1 & $\Delta$ Err. \\
 \midrule
WSJ Test & 91.47 & 92.13 & -7.7\% & 95.71 & -49.7\%\\
Brown All & 85.60 & 86.78 & -8.2\% & 93.53 & -55.0\%\\
Genia All & 80.31 & 81.64 & -6.8\% & 87.75 & -37.8\%\\
EWT All & 79.07 & 80.50 & -6.8\% & 89.27 & -48.7\%\\
\midrule
CTB Test & 83.67 & 85.69 & -12.4\% & 91.81 & -49.9\%\\
 B.\ News & 77.83 & 81.64 & -17.2\% & 88.41 & -47.7\%\\
Forums & 75.71 & 79.44 & -15.4\% & 87.04 & -46.6\%\\
 Blogs & 74.74 & 78.21 & -13.7\% & 84.29 & -37.8\%\\
 B.\ Conv.\ & 67.69 & 70.34 & \phantom{0}-8.2\% & 75.88 & -25.3\%\\
    \bottomrule
    \end{tabular}
    }
    \caption{\label{tab:inorder_pretraining}Performance of the In-Order parser, comparing using no pre-trained representations (first column), word embeddings, and BERT, on English (top) and Chinese (bottom) corpora. $\Delta$ Err.\ shows change in F1 error relative to the base parser (without pretraining). For both pre-training methods, error reduction is not typically greater out-of-domain than in-domain.}
    \vspace{-1em}
\end{table}

\subsection{Word embeddings}
We use the same pre-trained word embeddings as the original In-Order English and Chinese parsers,%
\footnote{
\href{https://github.com/LeonCrashCode/InOrderParser}{https://github.com/LeonCrashCode/InOrderParser}
}
trained on English and Chinese Gigaword \cite{parker2011english} respectively. %
Table~\ref{tab:inorder_pretraining} compares models without (In-Order column) to models with embeddings (+Embeddings),
showing that embeddings give comparable error reductions both in-domain (the WSJ Test and CTB Test rows) and out-of-domain (the other rows).

\begin{table}
    \centering
    \footnotesize
    \begin{tabular}{r|c|cc}
    \toprule
     &  Chart &  \multicolumn{2}{c}{+BERT} \\
 &  F1 &  F1 & $\Delta$ Err. \\
 \midrule
WSJ Test & 93.27 & 95.64 & -35.2\%\\
Brown All & 88.04 & 93.10 & -42.3\%\\
Genia All & 82.68 & 87.54 & -28.1\%\\
EWT All & 82.22 & 88.72 & -36.6\%\\
\bottomrule
\end{tabular}
    \caption{\label{tab:chart_pretraining} Performance of the Chart parser on English, comparing using no pretrained representations to using BERT. $\Delta$ Err.\ shows change in F1 error relative to the base parser. BERT does not generally provide a larger error reduction out-of-domain than in-domain.}
    \vspace{-1em}
\end{table}

\subsection{BERT}
For the Chart parser, we compare the base neural model (Sec.~\ref{sec:setup} and \ref{sec:neural_generalization}) to a model that uses a pre-trained BERT encoder \cite{kitaev2018multilingual}, using the publicly-released code\footnote{
\href{https://github.com/nikitakit/self-attentive-parser}{https://github.com/nikitakit/self-attentive-parser}
} to train and evaluate both models. %

For the In-Order parser, we introduce a novel integration of a BERT encoder with the parser's structured tree decoder. These architectures represent the best-performing types of encoder and decoder, respectively, from past work on constituency parsing, but have not been previously combined. We replace the word embeddings and predicted part-of-speech tags in the In-Order parser's stack and buffer representations with BERT's contextual embeddings. See Appendix~\ref{sec:bert_integration} for details on the architecture. Code and trained models for this system are publicly available.\footnote{
\href{https://github.com/dpfried/rnng-bert}{https://github.com/dpfried/rnng-bert}} %

Both the Chart and In-Order parsers are trained in the same way: the parameters of the BERT encoder (BERT\textsubscript{LARGE, Uncased} English or BERT\textsubscript{BASE} Chinese) are fine-tuned during training on the treebank data, along with the parameters of the parser's decoder. See Appendix~\ref{sec:bert_optimization} for details.

Results for the In-Order parser are shown in the +BERT section of Table~\ref{tab:inorder_pretraining}, and results for the chart parser are shown in Table~\ref{tab:chart_pretraining}. BERT is effective across domains, providing between 25\% and 55\% error reduction over the base neural parsers.
However, as for word embeddings, the pre-trained BERT representations do not generally provide a larger error reduction in out-of-domain settings than in in-domain  (although a possible confound is that the BERT model is fine-tuned on the relatively small amount of in-domain treebank data, along with the other parser parameters).

For English, error reduction from BERT is comparable between WSJ and EWT, largest on Brown, and smallest on Genia, which may indicate a dependence on the similarity between the out-of-domain treebank and the pre-training corpus.\footnote{BERT is pre-trained on books and Wikipedia; Genia consists of biomedical text.} %
For Chinese, the relative error reduction from BERT is largest on the in-domain CTB Test corpus.

\section{Can structure improve performance?}
\label{sec:structure}

\begin{table}[t]
\centering
\footnotesize
\begin{tabular}{r|cc|cc}
\toprule
 & \multicolumn{2}{c|}{F1} & \multicolumn{2}{c}{Exact Match} \\
 & Chart & In-Order & Chart & In-Order \\
 & +BERT & +BERT & +BERT & +BERT \\
 \midrule
WSJ Test & 95.64 & 95.71 & 55.11 & 57.05\\
Brown All & 93.10 & 93.54 & 49.23 & 51.98\\
EWT All & 88.72 & 89.27 & 41.83 & 43.98\\
Genia All & 87.54 & 87.75 & 17.46 & 18.03\\
\midrule
CTB Test & 92.14 & 91.81 & 44.42 & 44.94\\
B. News & 88.21 & 88.41 & 15.91 & 17.29\\
Forums & 86.72 & 87.04 & 20.00 & 21.95\\
Blogs & 84.28 & 84.29 & 17.14 & 18.85\\
B. Conv. & 76.35 & 75.88 & 17.24 & 18.99\\
\bottomrule
    \end{tabular}
\caption{\label{tab:exact_match}F1 and exact match accuracies comparing the Chart (unstructured) and In-Order (structured) parsers with BERT pretraining on English (top) and Chinese (bottom) corpora. 
\vspace{-1em}
}
    \label{tab:my_label}
\end{table}

When using BERT encoder representations, the Chart parser (with its unstructured decoder) and In-Order parser (with its conditioning on a representation of previously-constructed structure) obtain roughly comparable F1 (shown in the first two columns of Table~\ref{tab:exact_match}), with In-Order better on seven out of nine corpora but often by slight margins. %
However, these aggregate F1 scores decompose along the structure of the tree, and are dominated by the short spans which make up the bulk of any treebank. Structured-conditional prediction may plausibly be most useful for predicting larger portions of the tree, measurable in exact match accuracies and in F1 on longer-length spans (containing more substructure).

\begin{figure*}[t!]
  \centering
 
  \begin{subfigure}[b]{0.14\textwidth}
    \centering
    \includegraphics[scale=0.32]{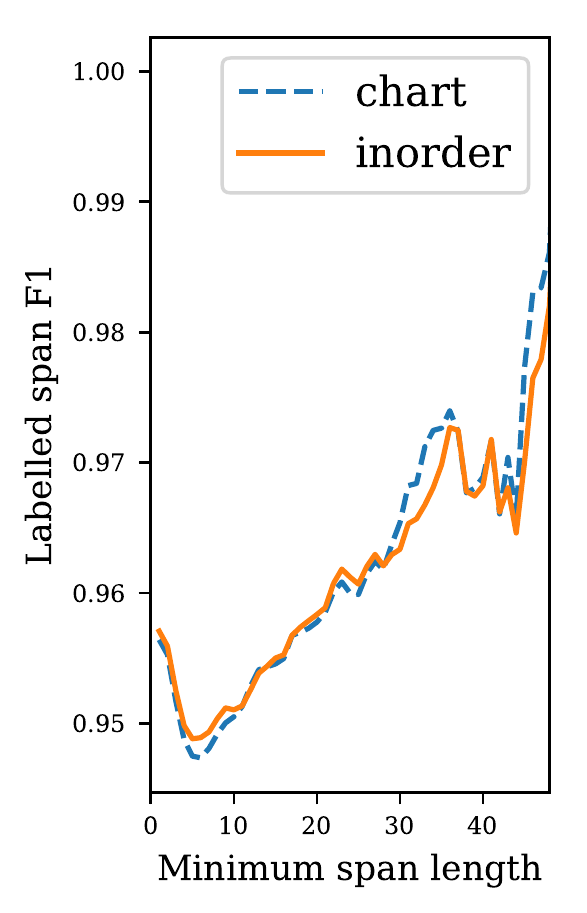}
    \caption{WSJ Test}
  \end{subfigure}
  \begin{subfigure}[b]{0.27\textwidth}
    \centering
    \includegraphics[scale=0.32]{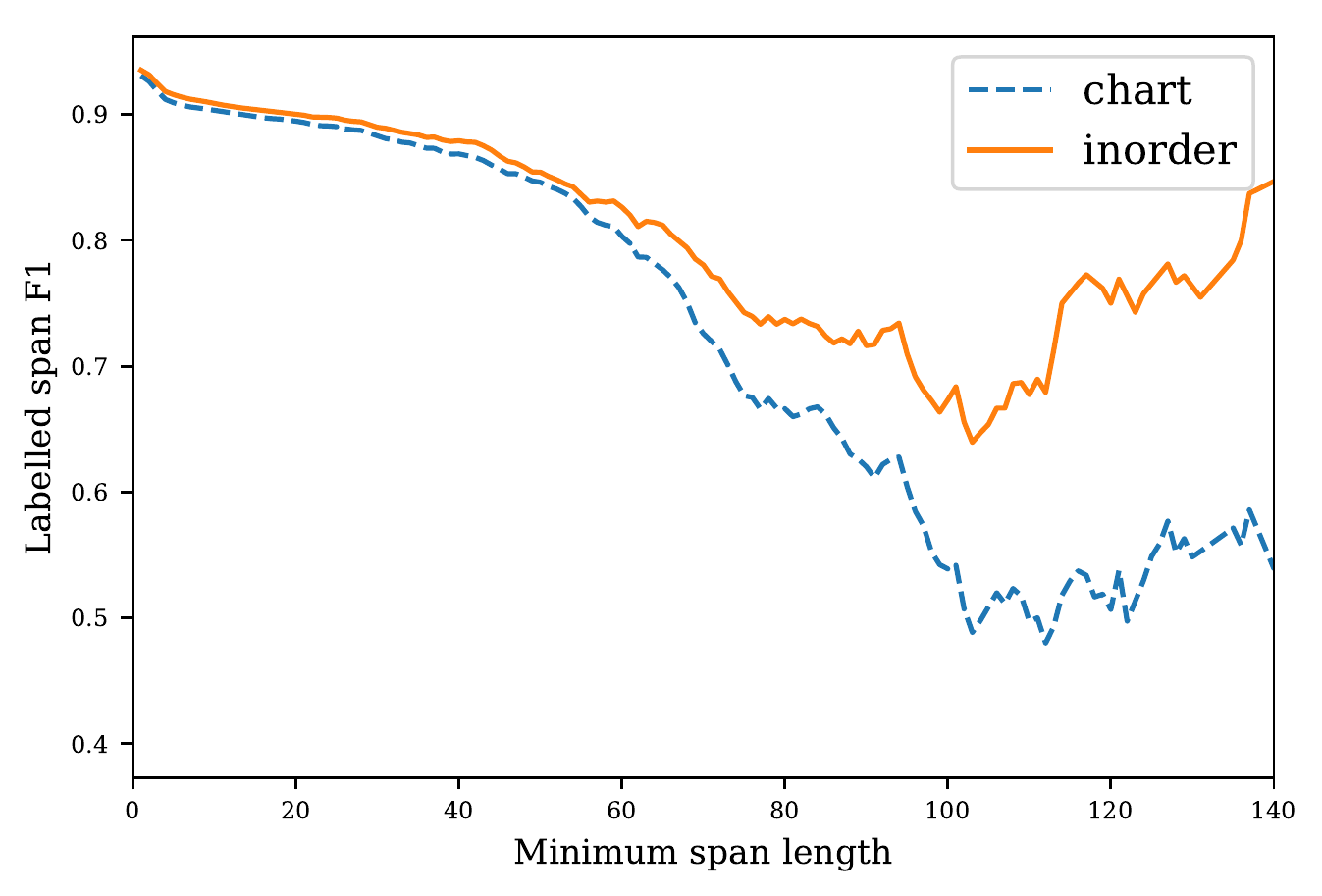}
    \caption{Brown All}
  \end{subfigure}
  \begin{subfigure}[b]{0.27\textwidth}
    \centering
    \includegraphics[scale=0.32]{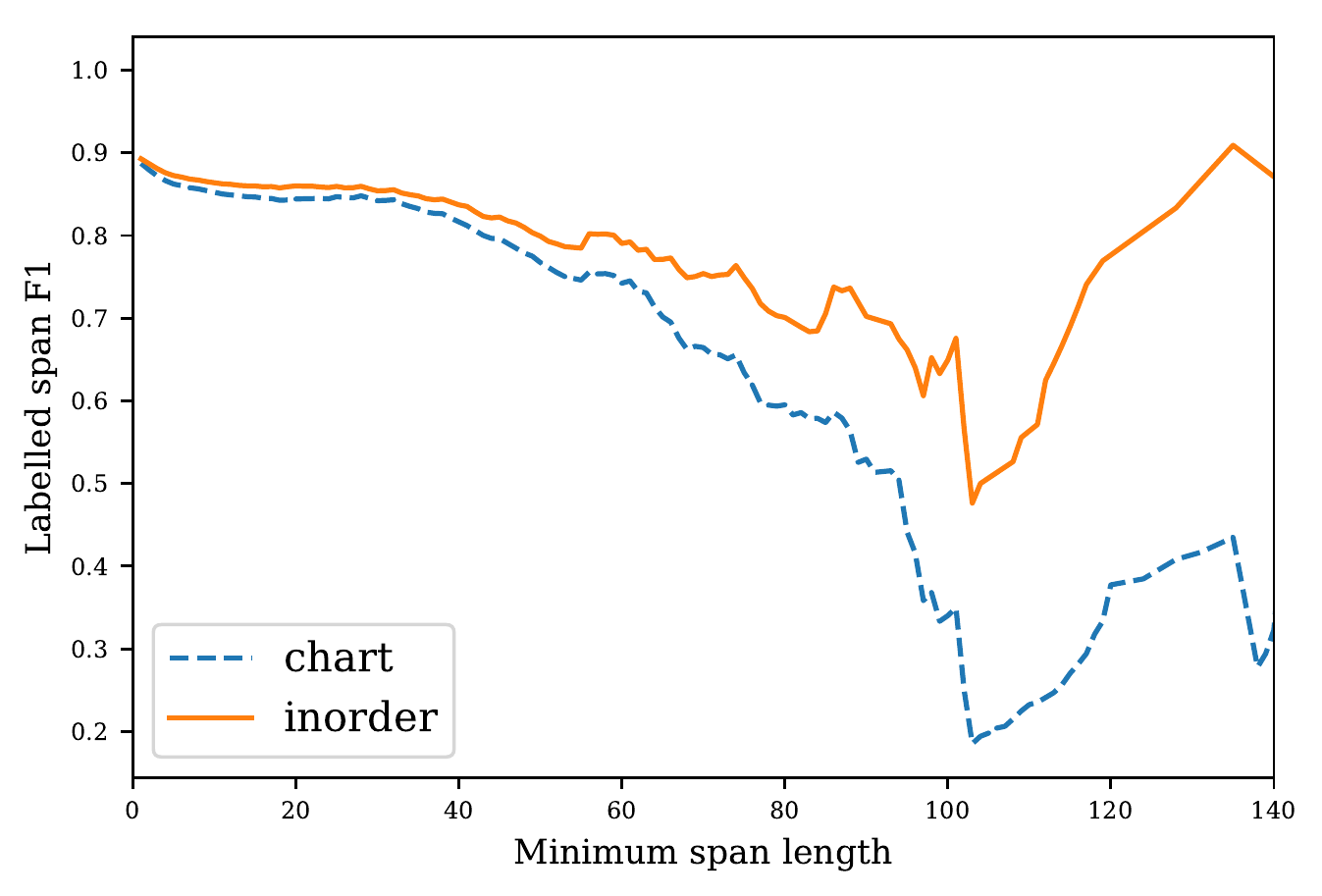}
    \caption{EWT All}
  \end{subfigure}
  \begin{subfigure}[b]{0.27\textwidth}
    \centering
    \includegraphics[scale=0.32]{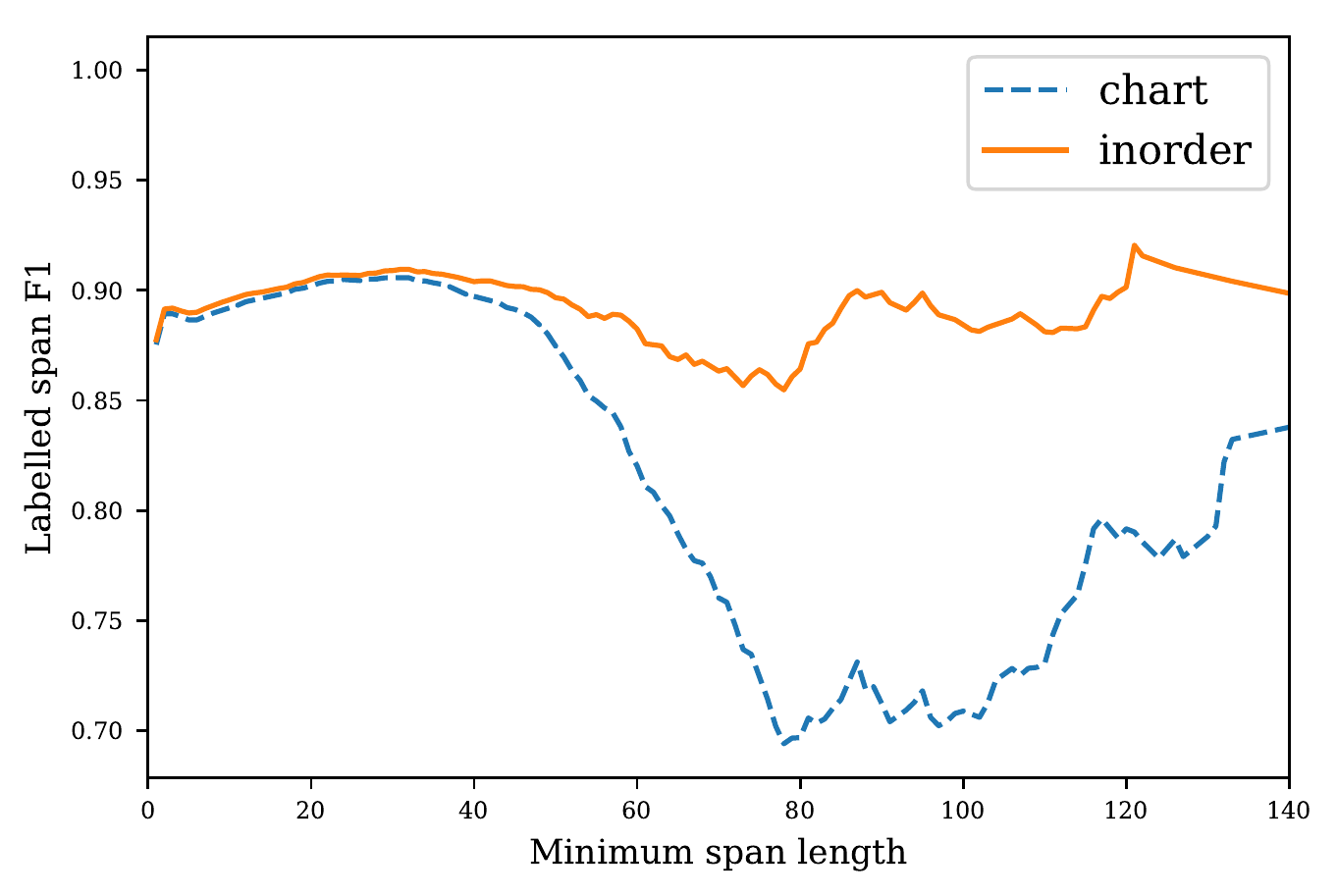}
    \caption{Genia All}
  \end{subfigure}
  \caption{
  \label{fig:length}
  Labelled bracketing F1 versus minimum span length for the English corpora. F1 scores for the In-Order parser with BERT (orange) and the Chart parser with BERT (cyan) start to diverge for longer spans.}
\end{figure*}

First, we compare the tree-level exact match accuracies of the two parsers. %
In the last two columns of Table~\ref{tab:exact_match}, we see that the In-Order parser consistently achieves higher exact match than the Chart parser across domains (including the in-domain WSJ and CTB Test sets), with improvements ranging from 0.5 to 2.8 percentage absolute. In fact, for several corpora (Blogs and B. Conv) the In-Order parser outperforms the Chart parser on exact match despite having the same or lower F1. This suggests that conditioning on structure in the model induces a correlation between span-level decisions that becomes most apparent when using a 
metric defined on the entire structure.

Second, we compare the performance of the two parsers on longer spans of text. Figure~\ref{fig:length} plots F1 by minimum span length for the In-Order and Chart parsers with BERT encoders on the English treebanks.
Across datasets, the improvement of the In-Order parser is slight when computing F1 across all spans in the dataset ($x=0$), but becomes pronounced when considering longer spans. %
This effect is not observed in the WSJ test set, which may be attributable  to its lack of sufficiently many long spans for us to observe a similar effect there.
The curves start to diverge at span lengths of around 30--40 words, longer than the median length of a sentence in the WSJ (23 words).

\section{Discussion}

\begin{table}
\centering
\footnotesize
\begin{tabular}{r|c|c|c}
\toprule
& & Chart & In-Order \\
& prior work & +BERT & +BERT \\
\midrule
Brown Test & 87.7 (C+'15) & 93.16  & 93.66 \\
Genia Test & 79.4 (C+'15) & 86.11 & 86.45 \\
EWT Test & \hspace{0.2em}83.5 (L+'12) & 89.13 & 89.62 \\
\bottomrule
\end{tabular}
\caption{\label{tab:comparison}Comparison of F1 scores for neural models with BERT pretraining to past state-of-the art results on transfer to the out-of-domain treebanks: (C+'15: \citealt{choe2015fusion}, L+'12: \citealt{leroux2012dcu}).\footnotemark\hspace{0.5mm}
EWT scores are averaged across the 3 SANCL'12 test sets, as reported by \citet{petrov2012overview}. %
}
\end{table}
\footnotetext{Although the F1 scores obtained here are higher than the zero-shot transfer results of \citet{joshi2018extending} on the Brown and Genia corpora due to the use of improved encoder (BERT) and decoder (self-attentive Chart and In-Order) models, we note the results are not directly comparable due to the use of different sections of the corpora for evaluation.}

Neural parsers generalize surprisingly well, and are able to draw benefits both from pre-trained language representations and structured output prediction. These properties %
allow single-model parsers to surpass previous state-of-the-art systems  on out-of-domain generalization (Table \ref{tab:comparison}).
We note that these systems from prior work \cite{choe2015fusion,petrov2012overview,leroux2012dcu} use additional ensembling or self-training techniques, which have also been shown to be compatible with neural constituency parsers \cite{dyer2016rnng,choe2016parsing,fried2017combination,kitaev2018multilingual} and may provide benefits orthogonal to the pre-trained representations and structured models we analyze here. 
Encouragingly, parser improvements on the WSJ and CTB treebanks still transfer out-of-domain, indicating that improving results on these benchmarks may still continue to yield benefits in broader domains.

\section*{Acknowledgements}
This research was supported by DARPA through the XAI program, as well as by a Tencent AI Lab fellowship to the first author. This research used the Savio computational cluster provided by the Berkeley Research Computing program at the University of California, Berkeley.

\bibliography{acl2019}
\bibliographystyle{acl_natbib}

\clearpage

\appendix

\section{Appendix}
\subsection{Integrating BERT into the In-Order Parser}
\label{sec:bert_integration}
In this section we describe our integration of the BERT encoder into the In-Order parser decoder. We refer to the original In-Order \cite{liu2017inorder} and BERT \cite{devlin2018bert} papers for full details about the model architectures, only describing the modifications we make at the interface between the two.
Code and pre-trained models for this integrated parser are publicly available.\footnote{\href{https://github.com/dpfried/rnng-bert}{https://github.com/dpfried/rnng-bert}}

BERT divides each word in an input sentence into one or more subword units and produces a contextual representation for each subword unit using a self-attentive architecture \cite{devlin2018bert}. Following the implementation of \citet{kitaev2018multilingual} for the Chart parser, we take the contextual representation vector for the last subword unit in each word $w_i$ as the word's representation, $e_{w_i}$, replacing the (non-contextual) word and POS tag vectors used in the original In-Order parser. We use a learned linear projection to scale $e_{w_i}$ to a vector $x_i$ of size 128 (compare with section 4.1 of \citet{liu2017inorder}).

These contextual word representations $x_i$ enter into the In-Order parser's decoder in two positions: the stack (representing the parse tree as constructed so far) and the buffer (representing the remainder of the sentence to be parsed). We retain the stack representation, but omit the LSTM which the original In-Order work uses to summarize the words remaining on the buffer. We instead use the representation $x_i$ as the buffer summary for the word $i$ when $i$ is word at the front of the buffer (the next word in the sentence to be processed). In early experiments we found that removing the LSTM summary of the buffer in this manner had no consistent effect on performance, indicating that the BERT contextual vectors already sufficiently aggregate information about the input sentence so that an additional LSTM provides no further benefit.

We pass values and gradients between the DyNet \cite{neubig2017dynet} implementation of the In-Order parser and the Tensorflow \cite{abadi2016tensorflow} implementation of BERT using the Tensorflow C++ API.

\subsection{BERT Optimization Settings}
\label{sec:bert_optimization}

We train the In-Order parser with BERT following the optimization procedure used in \citet{kitaev2018multilingual}'s publicly-released implementation of the BERT Chart parser: training with mini-batches of size 32 using the Adam optimizer \cite{kingma2014adam}; halving the base learning rates for Adam whenever 2 epochs of training pass without improved F1 on the development set, and using a warmup period for the BERT learning rate. For the In-Order parser, we use initial Adam learning rates of \num{2e-5} for the BERT encoder parameters and \num{1e-3} for the In-Order decoder parameters, $\beta_1 = 0.9$, $\beta_2=0.999$, and a BERT learning rate warmup period of 160 updates.

\end{document}